\documentclass[journal,twoside,web]{ieeecolor}
\usepackage{generic}
\usepackage{cite}
\usepackage{amsmath,amssymb,amsfonts}
\usepackage{bbm}
\usepackage{graphicx}
\usepackage{hyperref}
\usepackage{caption,subcaption}
\usepackage{multirow}
\usepackage{cleveref}
\usepackage{comment}
\usepackage{hhline}
\usepackage{todonotes}
\usepackage{graphicx}
\usepackage{caption,subcaption}
\usepackage{siunitx}
\usepackage{booktabs}
\usepackage{csvsimple}
\usepackage{datatool}
\usepackage{array}

\def\BibTeX{{\rm B\kern-.05em{\sc i\kern-.025em b}\kern-.08em
    T\kern-.1667em\lower.7ex\hbox{E}\kern-.125emX}}
\begin{document}

\title{Deep Learning for ECG Analysis: Benchmarks and Insights from PTB-XL}
\author{Nils Strodthoff*, Patrick Wagner*, Tobias Schaeffter and Wojciech Samek, \IEEEmembership{Member, IEEE}
\thanks{This work was supported by the Bundesministerium für Bildung und Forschung through the BIFOLD - Berlin Institute for the Foundations of Learning and Data (ref. 01IS18025A and ref 01IS18037A) and by the EMPIR project 18HLT07 MedalCare.}
\thanks{Nils Strodthoff, Patrick Wagner and Wojciech Samek are with Fraunhofer Heinrich Hertz Institute, 10587 Berlin, Germany (email: firstname.lastname@hhi.fraunhofer.de). Tobias Schaeffter is with Physikalisch-Technische Bundesanstalt, 10587 Berlin, Germany, Technical University Berlin, 10587 Berlin, Germany, and King's College London, London WC2R 2LS, UK (email: tobias.schaeffter@ptb.de).}
\thanks{*Both authors contributed equally to this work (Corresponding authors: Nils Strodthoff, Wojciech Samek).}}

\maketitle

\bstctlcite{BSTcontrol}

\maketitle

\begin{abstract}
Electrocardiography is a very common, non-invasive diagnostic procedure and its interpretation is increasingly supported by automatic interpretation algorithms. The progress in the field of automatic ECG interpretation has up to now been hampered by a lack of appropriate datasets for training as well as a lack of well-defined evaluation procedures to ensure comparability of different algorithms. To alleviate these issues, we put forward first benchmarking results for the recently published, freely accessible PTB-XL dataset, covering a variety of tasks from different ECG statement prediction tasks over age and gender prediction to signal quality assessment. We find that convolutional neural networks, in particular resnet- and inception-based architectures, show the strongest performance across all tasks outperforming feature-based algorithms by a large margin. These results are complemented by deeper insights into the classification algorithm in terms of hidden stratification, model uncertainty and an exploratory interpretability analysis. We also put forward benchmarking results for the ICBEB2018 challenge ECG dataset and discuss prospects of transfer learning using classifiers pretrained on PTB-XL. With this resource, we aim to establish the PTB-XL dataset as a resource for structured benchmarking of ECG analysis algorithms and encourage other researchers in the field to join these efforts.
\end{abstract}

\begin{IEEEkeywords}
    Decision support systems, Electrocardiography, Machine learning algorithms
    \end{IEEEkeywords}

\section{Introduction}
\label{sec:intro}
\newcommand{\percHighQuality}{$77.01$ }
\newcommand{\percGerman}{$70.89$ }
\newcommand{\percEnglish}{$27.9$ }
\newcommand{\percSwedish}{$1.21$ }
\newcommand{\percValBy}{$56.9$ }
\newcommand{\percValByHuman}{$73.7$ }
\newcommand{\percNotValByHuman}{$26.3$ }
\newcommand{\percExtraBeats}{$8.95$ }
\newcommand{\percDrift}{$7.36$ }
\newcommand{\percNoise}{$14.94$ }
\newcommand{\percBurst}{$2.81$ }
\newcommand{\percProblems}{$0.14$ }
\newcommand{\percPacemaker}{$1.34$ }
\newcommand{\percAutoInit}{$31.2$ }
\newcommand{\percHeight}{$31.98$ }
\newcommand{\percWeight}{$43.18$ }
\newcommand{\percPosition}{$61.05$ }
\newcommand{\numPatientsWithMutliple}{$2127$ }
\newcommand{\numSamples}{$21837$ }
\newcommand{\numPatients}{$18885$ }
\newcommand{\minDate}{1989/10/31 }
\newcommand{\maxDate}{1996/6/17 }
\newcommand{\sexMale}{$52$ }
\newcommand{\sexFemale}{$48$ }
\newcommand{\uniquesites}{$51$ }
\newcommand{\uniqueDoctors}{$12$ }
\newcommand{\uniqueNurses}{$12$ }
\newcommand{\uniqueDevices}{$11$ }
\newcommand{\numTotalLabels}{$71$ }
\newcommand{\numDiags}{$44$ }
\newcommand{\numForms}{$19$ }
\newcommand{\numRhythms}{$12$ }
\newcommand{\numInterDiagsForms}{$4$ }
\newcommand{\numInterFormsRhythms}{$0$ }
\newcommand{\traceOnly}{$1.67$ }
\newcommand{\percManually}{$67.13$ }
\newcommand{\percManuallyValidated}{$50.79$ }
\newcommand{\percAuto}{$31.2$ }
\newcommand{\percAutoValidated}{$4.45$ }
\newcommand{\percAutoNotValidated}{$26.75$ }
\newcommand{\ageMedian}{$62$ }
\newcommand{\ageIQR}{$22$ }
\newcommand{\weightMedian}{$70$ }
\newcommand{\weightIQR}{$20$ }
\newcommand{\heightMedian}{$166$ }
\newcommand{\heightIQR}{$14$ }
\newcommand{\minAge}{$0$ }
\newcommand{\maxAge}{$95$ }
\newcommand{\percValByHumanValidatedSet}{$56.9$ }
\newcommand{\percValByHumanSecondOpinionSet}{$0.62$ }
\newcommand{\percValByHumanNoAuto}{$16.18$ }
\newcommand{\percNotValByHumanAuto}{$26.3$ }

\IEEEPARstart{C}{ardiovascular} diseases (CVDs) rank among diseases of highest mortality \cite{wilkins2017statistics} and were in this respect only recently surpassed by cancer in high-income countries \cite{Dagenais2019}. Electrocardiography (ECG) is a non-invasive tool to assess the general cardiac condition of a patient and is therefore as first-in-line examination for diagnosis of CVD. In the US, during about 5\% of the office visits an ECG was ordered or provided \cite{NACMS2016}. In spite of these numbers, ECG interpretation remains a difficult task even for cardiologists \cite{Salerno2003} but even more so for residents, general practioners \cite{Salerno2003,Fent2015} or doctors in the emergency room who have to interprete ECGs urgently. 
A second major application area that will even grow in importance in the future is the telemedicine, in particular the monitoring of Holter ECGs. In both of these exemplary cases medical personnel could profit from significant reliefs if they were supported by advanced decision support systems relying on automatic ECG interpretation algorithms.

During recent years, we have witnessed remarkable advances in automatic ECG interpretation algorithms. In particular, deep-learning-based approaches have reached or even surpassed cardiologist-level performance for selected subtasks \cite{Attia2016, hannun2019cardiologist, Attia2019AF, Attia2019CCD, Ribeiro2020} or enabled statements that were very difficult to make for cardiologists e.g.\ to accurately infer age and gender from the ECG \cite{Attia2019Age}. 
Due to the apparent simplicity and reduced dimensionality compared to imaging data, also the broader machine learning community has gained a lot of interest in ECG classification as documented by numerous research papers each year, see \cite{hong2019opportunities} for a recent review. 

We see deep learning algorithms in the domain of computer vision as a role model for the deep learning algorithms in the field of ECG analysis. The tremendous advances for example in the field of image recognition relied crucially on the availability of large datasets and the competitive environment of classification challenges with clear evaluation procedures. In reverse, we see these two aspects as two major issues that hamper the progress in algorithmic ECG analysis: First, open ECG datasets are typically very small \cite{Schlapfer2017} and existing large datasets remain inaccessible for the general public. This issue has been at least partially resolved by the publication of the PTB-XL dataset \cite{PTB-XLphysionet,Wagner2019PTBXL} hosted by PhysioNet \cite{physionet}, which provides a freely accessible ECG dataset of unprecedented size with predefined train-test splits based on stratified sampling. Second, the existing datasets typically provide only the raw data, but there exist no clearly defined benchmarking tasks with corresponding evaluation procedures. This severely restricts the comparability of different algorithms, as experimental details such as sample selection, train-test splits, evaluation metrics and score estimation can largely impact the final result. To address this second issue, we propose a range of different tasks showcasing the variability of the dataset ranging from the prediction of ECG statements over age and gender prediction to the assessment of signal quality. For these tasks, we present first benchmarking results for deep-learning-based time series classification algorithms. We use the ICBEB2018 dataset to illustrate the promising prospects of transfer learning especially in the small dataset regime establishing PTB-XL as a pretraining resource for generic ECG classifiers, very much like ImageNet \cite{deng2009imagenet} in the computer vision domain. 

Finally, assessing the quantitative accuracy is an important but by far not the only important aspect for decision support systems in the medical domain. To develop algorithms that create actual clinical impact, the topics of interpretability, robustness in a general sense and model uncertainty deserve particular attention. Such deeper insights, which go beyond benchmarking results, are discussed in the second part of the results section highlighting various promising directions for future research. In particular, we present a first evaluation of the diagnosis likelihood information provided within the dataset in comparison to model uncertainty as well as an outlook to possible applications of interpretability methods in the field.

\section{Materials \& methods}
\subsection{PTB-XL dataset}
\label{sec:PTBXL}
\begin{figure}[ht]
    \centering
    \includegraphics[width=.8\columnwidth]{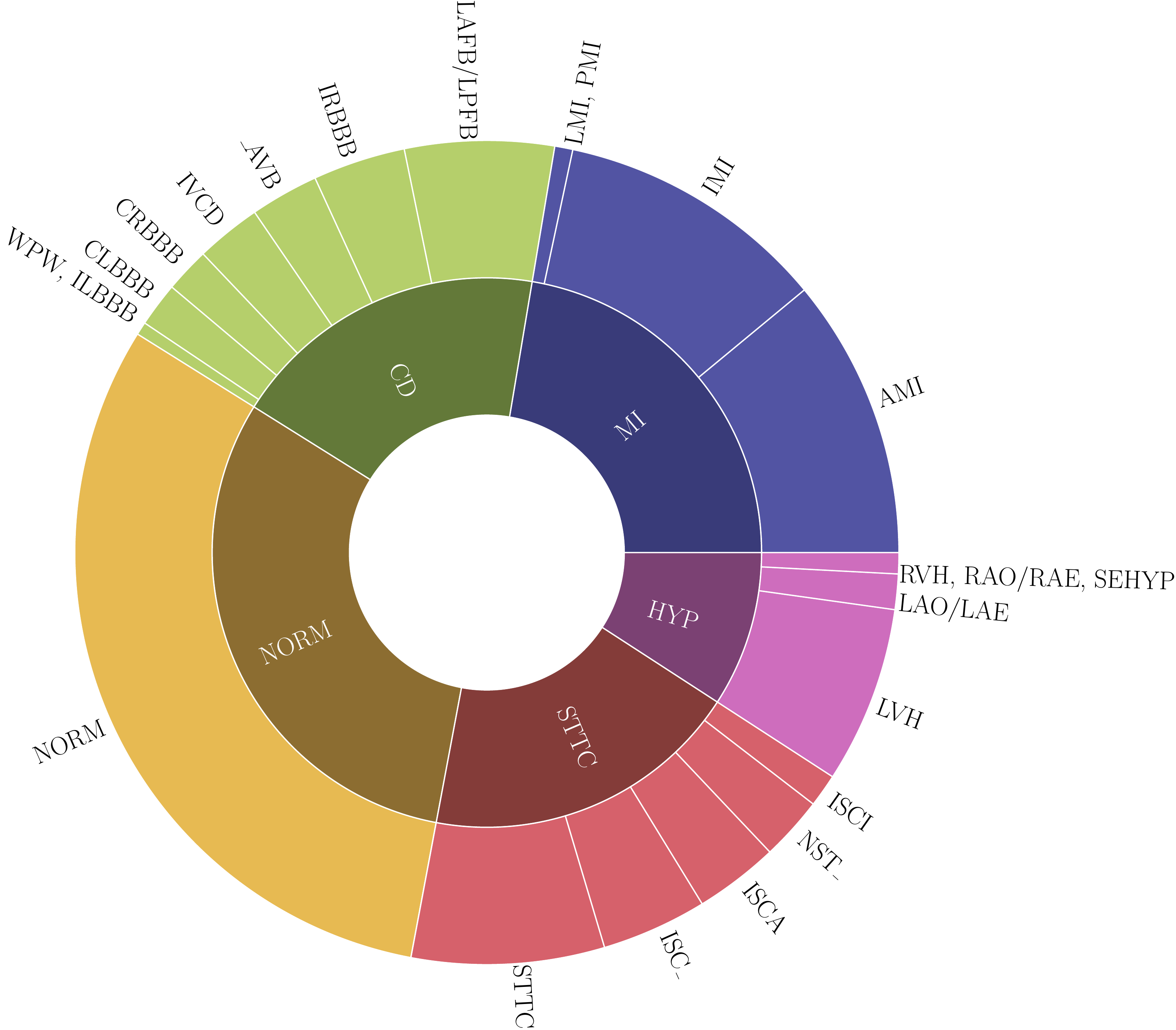}
    \caption{Summary of the PTB-XL dataset in terms of diagnostic super and subclasses where the size of area represents the fraction of samples (figure adapted from \cite{Wagner2019PTBXL}).}
    \label{fig:ptbxl_piechart}
\end{figure}

In this section, we briefly introduce the PTB-XL dataset \cite{Wagner2019PTBXL} that underlies most experiments presented below.
The PTB-XL dataset comprises \numSamples clinical 12-lead ECG records of 10 seconds length from \numPatients patients, where \sexMale\!\% were male and \sexFemale\!\% were female. The ECG statements used for annotation are conform to the SCP-ECG standard \cite{iso2009scpecg} and were assigned to three non-mutually exclusive categories \emph{diag.} (short for diagnostic), \emph{form} and \emph{rhythm}. In total, there are \numTotalLabels different statements, which decompose into \numDiags diagnostic, \numRhythms rhythm and \numForms form statements. Note that there are \numInterDiagsForms form statements that are also assigned to the set of diagnostic ECG statements. For diagnostic statements also a hierarchical organization into five coarse superclasses (\textit{NORM}: normal ECG, \textit{CD}: conduction disturbance, \textit{MI}: myocardial infarction, \textit{HYP}: hypertrophy and \textit{STTC}: ST/T changes) and 24 sub-classes is provided, see \Cref{fig:ptbxl_piechart}. For further details on the dataset and the annotation scheme, we refer the reader to the original publication \cite{Wagner2019PTBXL}. To illustrate the versatility of tasks that can be addressed within the dataset, we also incorporate the further metadata provided, namely demographic information such as age and gender or signal quality as assessed by a technical expert.

\subsection{Time series classification algorithms}
\label{sec:algos}
For benchmarking different classification algorithms, we focus on algorithms that operate on raw multivariate time series data. An alternative class of algorithms operates on derived or transformed features such as Fourier or Wavelet coefficients or handcrafted features extract from single beats after beat segmentation, see \cite{Maglaveras1998,Houssein2017} for review in the context of ECG classification and \cite{bagnall2016great} for (mostly univariate) time series classification in general. Deep learning approaches for time series classification are covered in a variety of recent, excellent reviews \cite{gamboa2017deep,fawaz2019deep,yannick2019deep}.

We evaluate adaptations of a range of different algorithms from the literature that can be broadly categorized as follows, see Appendix~\ref{appendix:details} for experimental details:
\begin{itemize}
  \item \textbf{convolutional neural networks}: 
  \begin{itemize}
    \item \textbf{standard}: fully convolutional \cite{wang2017time}, Deep4Net \cite{schirrmeister2017deep} 
    \item \textbf{resnet-based}: one-dimensional adaptations of standard resnets \cite{wang2017time,he2016deep}, wide resnets \cite{zagoruyko2016wide} and xresnets \cite{he2019bag}
    \item \textbf{inception-based}: InceptionTime \cite{fawaz2019dreamtime}
  \end{itemize}
  \item \textbf{recurrent neural networks}: LSTM \cite{hochreiter1997long}, GRU \cite{cho-etal-2014-learning}
  \item \textbf{baseline classifiers}:
  \begin{itemize}
    \item \textbf{feature-based:} Wavelet + shallow NN inspired by \cite{Sharma2017}
    \item \textbf{naive}: predicting the frequency of each term in the training set
  \end{itemize}
\end{itemize}
For reasons of clarity, we only report the performance for selected representatives including the best-performing method for each group. Typically the differences within the different groups are rather small. For completeness, the full results including all architectures are available in the accompanying code repository \cite{coderepo}.

To encourage future benchmarking on this dataset, we release our repository \cite{coderepo} used to produce the results presented below along with instructions on how to evaluate the performance of custom classifiers in this framework. Finally, we would like to stress that the deep learning models were trained on the original time series data without any further preprocessing such as removing baseline wander and/or filtering, which are commonly used in literature approaches but introduce further hyperparameters into the approach.

\subsection{Multi-label classification metrics}
\label{sec:metrics}
In this subsection, we review metrics for multi-label classification problems, see \cite{Zhang2014} for a review on multi-label classification metrics and algorithms.
Multi-label classification metrics can be categorized broadly as \emph{sample-centric} and \emph{label-centric} metrics. The main difference between metrics from both categories is the question whether to first aggregate the scores across labels and then across samples or vice versa. To obtain a comprehensive view of the classification performance, we pick one exemplary metric from each category as proposed on theoretical grounds by \cite{wu2017unified}. Here, we focus on metrics that can be evaluated based on soft classifier outputs, where no thresholding has been applied yet, as this allows to get a more complete picture of the discriminative power of a given classification algorithm. In addition, it disentangles the selection of an appropriate classifier from the issue of threshold optimization, that will anyway have to be adjusted to match the clinical requirements rather than to optimize a certain global target metric.

\paragraph{Term-centric metrics}
In general label-centric metrics are based on averages across class-specific metrics, which can further subdivided into micro- and macro-averages. In our setting, macro-averaging is preferred, since we expect class imbalance and do not want the score to be dominated by a few large classes. In addition, the distribution of pathologies in the dataset does not follow the natural distribution in the population but rather reflects the data collection process.
Averaging class-wise AUCs over all classes yields the term-centric macro AUC (henceforth abbreviated as AUC), which we will use as primary evaluation metric.

\paragraph{Sample-centric metrics}
Sample-centric evaluation metrics measure how accurately classification algorithms assign labels to a given sample, which is an information-retrieval point of view.
For the selection of sample-centric metrics, we follow the evaluation procedures over the course of to-date three CAFA classification challenges \cite{Zhou2019CAFA3}. The CAFA challenges address protein function prediction, which is also an inherent multi-label problem and shows strong structurally similarities to the task of ECG classification. For a given prediction $P_i(\tau)$ for given threshold $\tau\in[0,1]$ and corresponding ground-truth annotations $T_i$, we can define sample-centric precision $\text{pr}(\tau)$, recall/sensitivity $\text{rc}(\tau)$
\begin{align}
	\text{pr}(\tau)&= \frac{1}{N_\tau} \sum_{i\in \mathcal{N}_\tau} \frac{\sum_f \mathbbm{1}(f\in P_i(\tau) \wedge f\in T_i)}{\sum_f \mathbbm{1}(f\in P_i(\tau))}
	\,,\nonumber\\
	\text{rc}(\tau)&= \frac{1}{N}\sum_{i=1}^{N} \frac{\sum_f \mathbbm{1}(f\in P_i(\tau) \wedge f\in T_i)}{\sum_f \mathbbm{1}(f\in T_i)}\,,
\end{align}

where $\mathcal{N}_\tau=\{i\in{1,\ldots,N}| \sum_f \mathbbm{1}(f\in P_i(\tau))>0\}$ and $N_\tau=|\mathcal{N}_\tau|$. Here, we handle a possibly vanishing denominator when calculating the average precision in the same way as it is done in the CAFA challenges \cite{Jiang2016} by restricting the mean to the subset of samples with at least one prediction at the given threshold. Note that this procedure assumes a single threshold rather than class-dependent thresholds. 
We focus on Fmax as secondary performance metric, which was considered as main metric in the CAFA challenge. To this end, one defines a threshold-dependent $F_1$-score as the harmonic mean of precision and recall, i.e.\
\begin{align}
	F_1(\tau)=\frac{2\, \text{pr}(\tau) \text{rc}(\tau)}{\text{pr}(\tau)+\text{rc}(\tau)}\,.
\end{align}
To summarize $F_1(\tau)$ by a single number, the threshold is varied and the maximum score, from now on referred to as Fmax, is reported. As in the CAFA challenge, the threshold is optimized on the respective test set for each classification task and classifier under consideration. This procedure allow for a black-box evaluation just based on soft classifier outputs.
\section{Benchmarking results on PTB-XL and ICBEB2018}
\label{sec:exp}
PTB-XL comes with a variety of labels and further metadata. The presented experiments in this section serve two purposes: On the one hand, we provide first benchmarking results for future reference and, on the other hand, they illustrate the versatility of analyses that can be carried out based on the PTB-XL dataset. In \Cref{sec:quantitative}, we evaluate classifiers for different selections and granularities of ECG statements, which represents the core of analysis. It is complemented by \Cref{sec:icbeb}, where we validate our findings on the ICBEB2018 dataset and investigate aspects of transfer learning using PTB-XL for pretraining. Finally, we illustrate ways of leveraging further metadata within PTB-XL to construct age and gender prediction models, see \Cref{sec:exp_age_gender}, and to build signal quality assessment models based on the provided signal quality annotations, see \Cref{sec:sig_quality}.

\subsection{ECG statement prediction on PTB-XL}
\label{sec:quantitative}
We start by introducing, performing and evaluating all experiments that are directly related to ECG-statements, where we cover the three different major categories diagnostic \textit{diag.}, \textit{form} and \textit{rhythm} and level (\textit{sub-diag.} and \textit{super-diag.} as proposed in \cite{Wagner2019PTBXL}) resulting in different number of labels per experiment and per sample as can seen in \Cref{tab:n_per_ecg}. In the next step, we select only samples with at least one label in the given label selection. Our proposed evaluation as described in \Cref{sec:metrics} is applied the same way for each experiment, where we report the term-centric macro-averaged AUC and the sample-centric Fmax-score. 

\begin{table}
  \centering
  \caption{Number of ECG statments per sample for a given level.}
  \label{tab:n_per_ecg}
\begin{tabular}{lrrrrrr}
  \toprule
                   Level &   \# classes  & 0 &      1 &      2 &     3 &     $\geq$4 \\
  \midrule
              diag. &  44 &  407 &  15019 &   4242 &  1515 &   654 \\
              sub diag. &  24 &  407 &  16272 &   4079 &   920 &   159 \\
              super-diag. &  5 &  407 &  15239 &   4171 &  1439 &   581 \\
                    form &  19 & 12849 &   6693 &   1672 &   524 &    99 \\
                  rhythm &  12 &  771 &  20923 &    142 &     1 &     0 \\
                     all &  71 &    0 &    705 &  11247 &  5114 &  4771 \\
  \bottomrule
  \end{tabular}
  
\end{table}
\begin{table*}
  \centering
  \setlength\tabcolsep{3.5pt}
  \caption{Overall discriminative performance of ECG classification algorithms on PTB-XL. For each experiment and each metric the best mean performing model is highlighted in bold font. For all experiments, 95\% confidence intervals were calculating via bootstrapping on the test set, see Appendix~\ref{appendix:details} for notation and further details.}
  \label{tab:summary_scp_ecg}
\begin{tabular}{|l|ll|ll|ll|ll|ll|ll|}
\toprule
  & \multicolumn{2}{c|}{all}& \multicolumn{2}{c|}{diag.}& \multicolumn{2}{c|}{sub-diag.}& \multicolumn{2}{c|}{super-diag.}& \multicolumn{2}{c|}{form}& \multicolumn{2}{c|}{rhythm}\\ Method & AUC & Fmax & AUC & Fmax & AUC & Fmax & AUC & Fmax & AUC & Fmax & AUC & Fmax \\
\midrule
 lstm & .907(10) & .757(08) & .927(09) & .731(11) & .928(10) & .759(11) & .927(05) & .820(09) & .851(19) & .604(23) & .953(10) & .910(08) \\
 inception1d & \bfseries .925(08) & .762(08) & .931(10) & .737(11) & \bfseries .930(10) & .752(13) & .921(06) & .810(11) & \bfseries .899(21) & .621(26) & .953(11) & \bfseries .917(08) \\
 lstm\_bidir & .914(09) & .761(08) & .932(08) & .737(12) & .923(12) & .757(12) & .921(06) & .815(10) & .876(18) & .625(22) & .949(11) & .912(09) \\
 resnet1d\_wang & .919(08) & \bfseries .767(08) & .936(08) & \bfseries .741(13) & .928(10) & \bfseries .762(12) & \bfseries .930(06) & \bfseries .823(10) & .880(19) & .628(25) & .946(10) & .911(10) \\
 fcn\_wang & .918(08) & .757(08) & .926(11) & .735(13) & .927(11) & .756(12) & .925(06) & .817(12) & .869(15) & .615(23) & .931(09) & .898(11) \\
 Wavelet+NN & .849(13) & .690(10) & .855(16) & .634(16) & .859(17) & .660(13) & .874(08) & .734(11) & .757(32) & .542(27) & .890(25) & .869(10) \\
 xresnet1d101 & \bfseries .925(08) & .764(10) & \bfseries .937(08) & .736(12) & .929(13) & .760(12) & .928(06) & .815(12) & .896(12) & \bfseries .643(25) & \bfseries .957(15) & \bfseries .917(08) \\
 \hline ensemble & \bfseries .929(08) & \bfseries .769(08) & \bfseries .939(09) & \bfseries .743(12) & \bfseries .933(12) & \bfseries .766(12) & \bfseries .934(05) & \bfseries .825(12) & \bfseries .907(14) & \bfseries .644(24) & \bfseries .965(07) & .915(09) \\
 \hline naive & \itshape .500(00) & \itshape .557(12) & \itshape .500(00) & \itshape .440(18) & \itshape .500(00) & \itshape .440(21) & \itshape .500(00) & \itshape .448(10) & \itshape .500(00) & \itshape .365(17) & \itshape .500(00) & \itshape .797(15) \\
\bottomrule
\end{tabular}

\end{table*}
\begin{figure*}
  \centering
  \begin{subfigure}[b]{0.3\textwidth}
      \caption{diag.}
      \includegraphics[width=\textwidth]{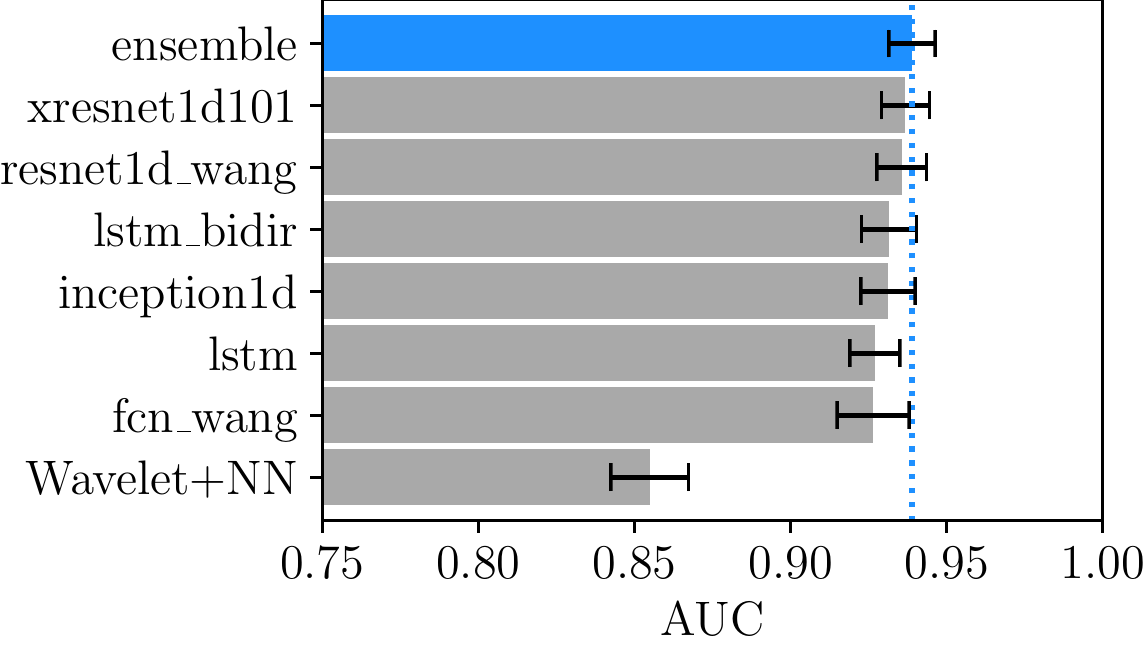}
      \label{fig:exp1}
  \end{subfigure}
  ~ 
  \begin{subfigure}[b]{0.3\textwidth}
    \caption{sub-diag.}
    \includegraphics[width=\textwidth]{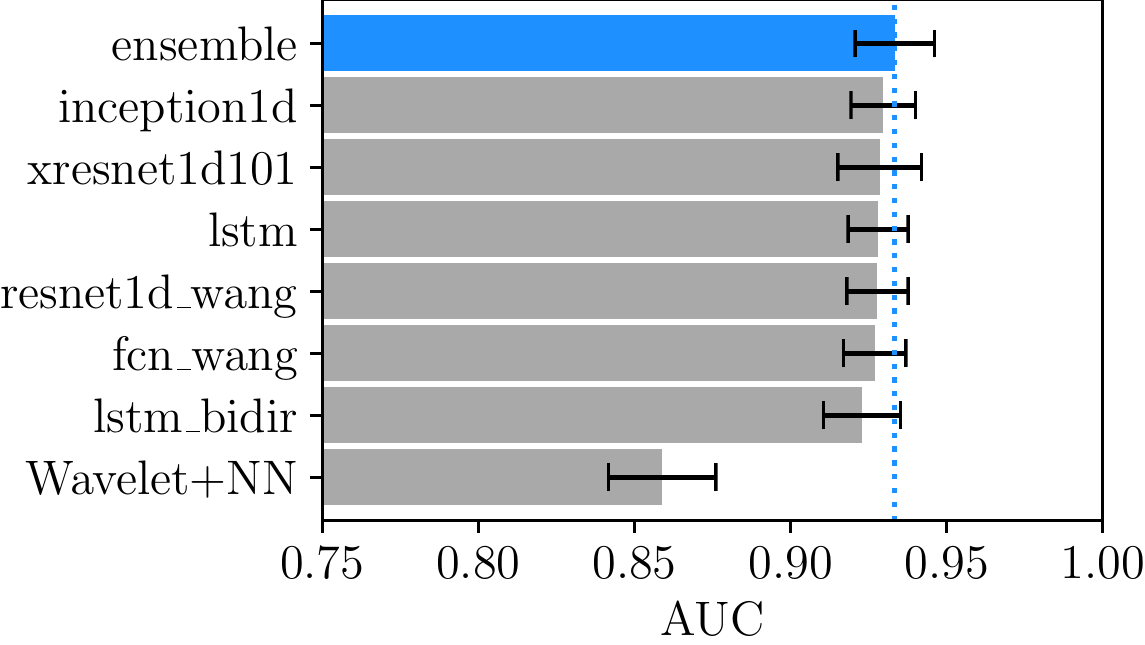}
    \label{fig:exp1.1}
  \end{subfigure}
  ~ 
  \begin{subfigure}[b]{0.3\textwidth}
    \caption{super-diag.}
    \includegraphics[width=\textwidth]{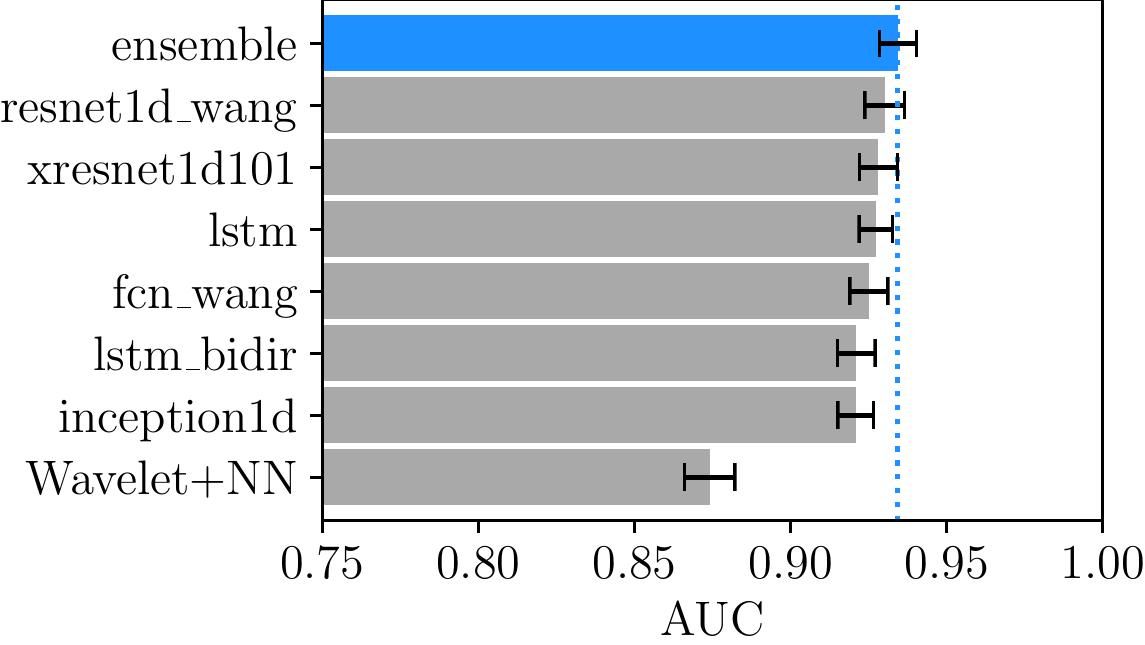}
    \label{fig:exp1.1.1}
  \end{subfigure}
  \\
  \vspace{-.5cm}
  \begin{subfigure}[b]{0.3\textwidth}
    \caption{all}
    \includegraphics[width=\textwidth]{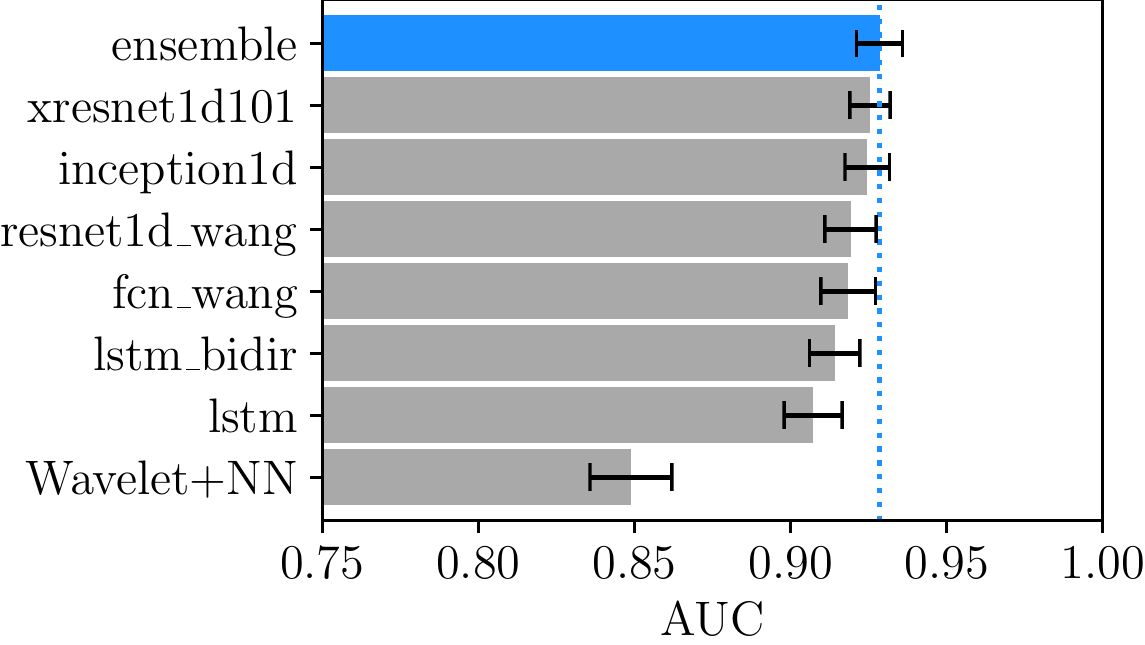}
    \label{fig:exp0}
  \end{subfigure}
  ~ 
  \begin{subfigure}[b]{0.3\textwidth}
    \caption{form}
    \includegraphics[width=\textwidth]{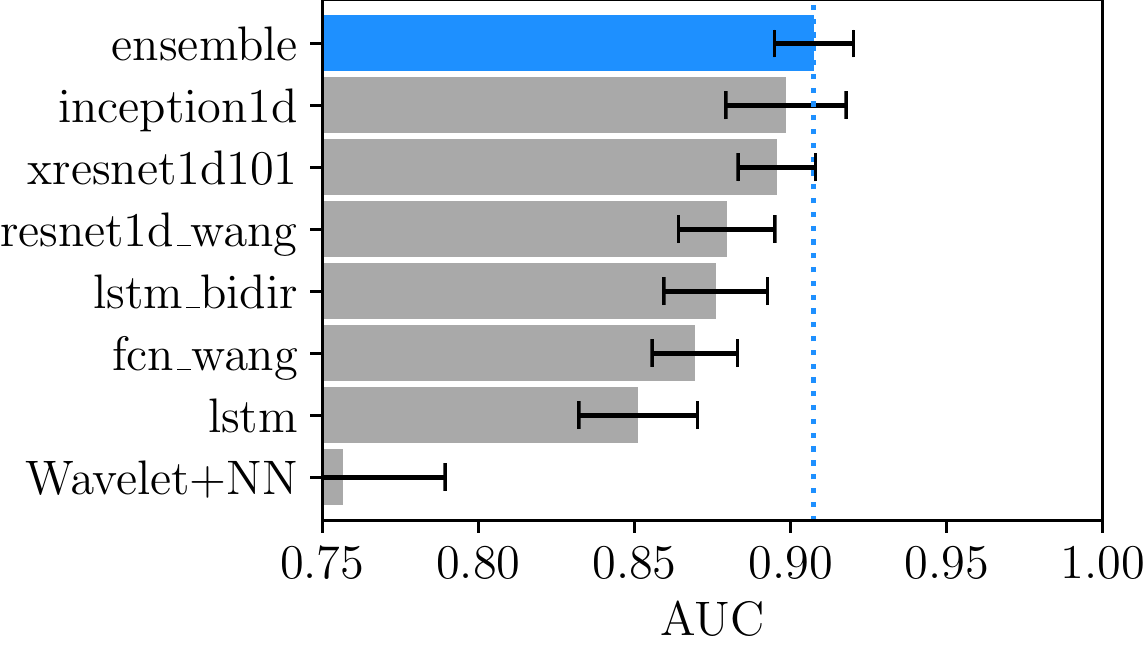}
    \label{fig:exp2}
  \end{subfigure}
  ~ 
  \begin{subfigure}[b]{0.3\textwidth}
    \caption{rhythm}
    \includegraphics[width=\textwidth]{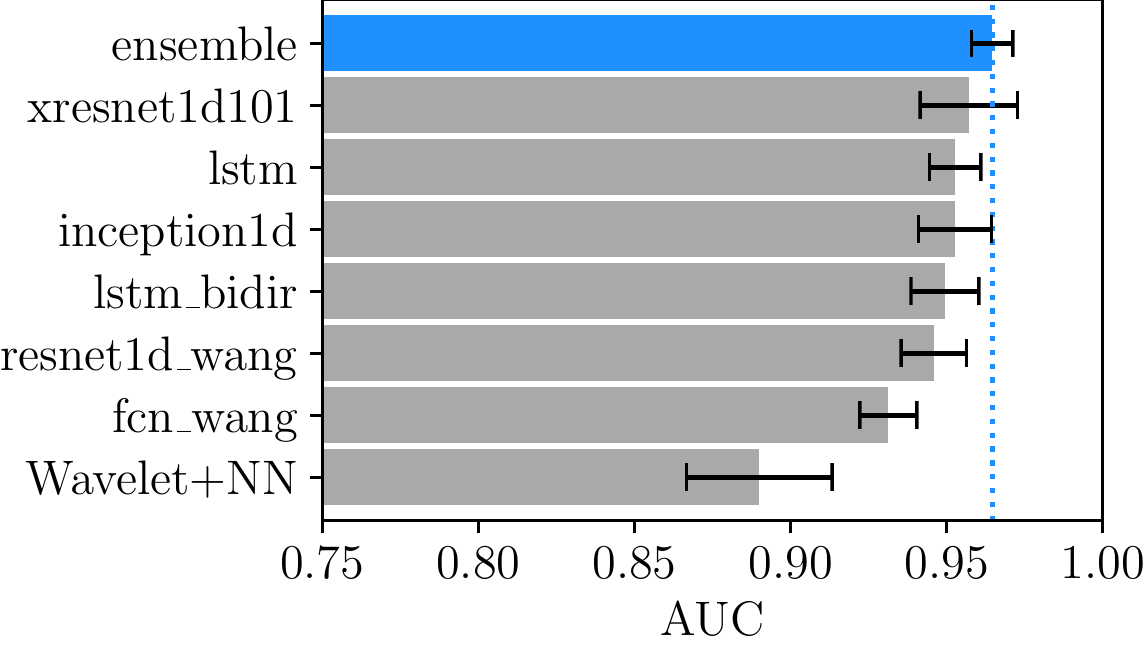}
    \label{fig:exp3}
  \end{subfigure}
  \vspace{-1\baselineskip}
  \caption{Graphical summary of experiments described in \Cref{sec:quantitative}. For comparability, the algorithms are ranked according to prediction performance in each category.}
  \label{fig:summary_scp_ecg2}
\end{figure*}
In \Cref{tab:summary_scp_ecg}, we report the results for all six experiments each applied to all models (as introduced in \Cref{sec:algos}), \Cref{fig:summary_scp_ecg2} shows the result for all six experiments using barplots with associated bootstrap confidence intervals, see Appendix~\ref{appendix:details} for details. In all six experiments, deep-learning-based methods show a high predictive performance. Interestingly, even though all models are optimized based on binary cross-entropy loss rather than on the target metrics directly, the ranking according to both sample-based and term-based metrics largely coincides across all algorithms, which is why we focus on macro AUC in the following. The best-performing resnet or inception-based models reach macro AUCs ranging from 0.89 in the \textit{form} category, over around 0.93 in the \textit{diagnostic} categories to 0.96 in the \textit{rhythm} category. 
These performance metrics can in principle used for a rudimentary assessment of the difficulty of the different prediction tasks. However, one has to keep in mind that for example the \textit{form} prediction task has a considerably smaller training set compared to the other experiments due to approximately 12k ECGs without any \textit{form} annotations. 

As first general observation upon investigating the different model performances in more detail, we find that resnet-architectures and inception-based architectures perform best across all experiments, but all convolutional architectures show a comparable performance level. In fact, the results of all convolutional models, up to very few exceptions, remain compatible within error bars. Recurrent architectures are consistently slightly less performant than their convolutional counterparts but, at least for diagnostic and rhythm statements, still competitive. The second general observation is that the performances of both convolutional as well as recurrent deep learning models turn out to be considerably stronger than the performance of the baseline algorithm operating on wavelet features. However, this statement has to be taken with caution, as the performance of feature-based classifiers is typically rather sensitive to details of feature selection choice of derived and details of the proprocessing procedure.

In addition to single-model-performance, we also report the performance of an ensemble formed by averaging the predictions of all considered models (except the naive model). As can be seen in \Cref{tab:summary_scp_ecg}, ensembling leads in many case to slight performance increases, but the best-performing single resnet or inception models always remain compatible with the ensemble result within error bars. The largest performance improvement of the ensemble model compared to single model performance is observed in the \textit{rhythm} category, where the ensemble model outperforms all convolutional models except for xresnet1d101 and inception1d (as can be seen in \Cref{fig:exp3}). The ensemble results are only supposed to serve as rough orientation as the focus of this work is on single-model performance.

As a final remark, throughout this paper we use the recommended train-test splits provided by PTB-XL \cite{Wagner2019PTBXL}, which consider patient assignments and use input data at a sampling frequency of 100~Hz. Deviations from this setup are investigated in Appendix~\ref{sec:misconceptions}.
\subsection{ECG statement prediction on ICBEB2018 and transfer learning}
\label{sec:icbeb}
\begin{table}[t]
  \centering
  \caption{Classification performance on the ICBEB2018 dataset. In addition to sample-centric Fmax and term-centric macro-AUC, we also report the term-centric $F_{\beta=2}$ and $G_{\beta=2}$ to be used in the PhysioNet/CinC challenge 2020.}
  \label{tab:icbeb}
\begin{tabular}{lllll}
\toprule
 Method & AUC & Fmax & $F_{\beta=2}$ & $G_{\beta=2}$ \\
\midrule
lstm & 0.964(07) & 0.827(18) & 0.790(27) & 0.561(34) \\
lstm\_bidir & 0.959(09) & 0.838(18) & 0.796(27) & 0.573(34) \\
xresnet1d101 & \bfseries 0.974(05) & \bfseries 0.855(20) & \bfseries 0.819(28) & \bfseries 0.602(44) \\
resnet1d\_wang & 0.969(08) & 0.849(20) & 0.803(30) & 0.586(41) \\
fcn\_wang & 0.957(07) & 0.824(23) & 0.787(31) & 0.563(38) \\
inception1d & 0.963(07) & 0.846(20) & 0.807(29) & 0.594(38) \\
Wavelet+NN & 0.905(12) & 0.701(22) & 0.665(35) & 0.405(33) \\
\hline naive & 0.500(00) & 0.289(21) & 0.368(06) & 0.115(00) \\
\hline ensemble & \itshape \bfseries 0.975(04) & \itshape \bfseries 0.863(17) & \itshape \bfseries 0.819(28) & \itshape \bfseries 0.608(37) \\
\bottomrule
\end{tabular}

\end{table}

\begin{figure}[ht]
 \centering
 \includegraphics[width=.8\columnwidth]{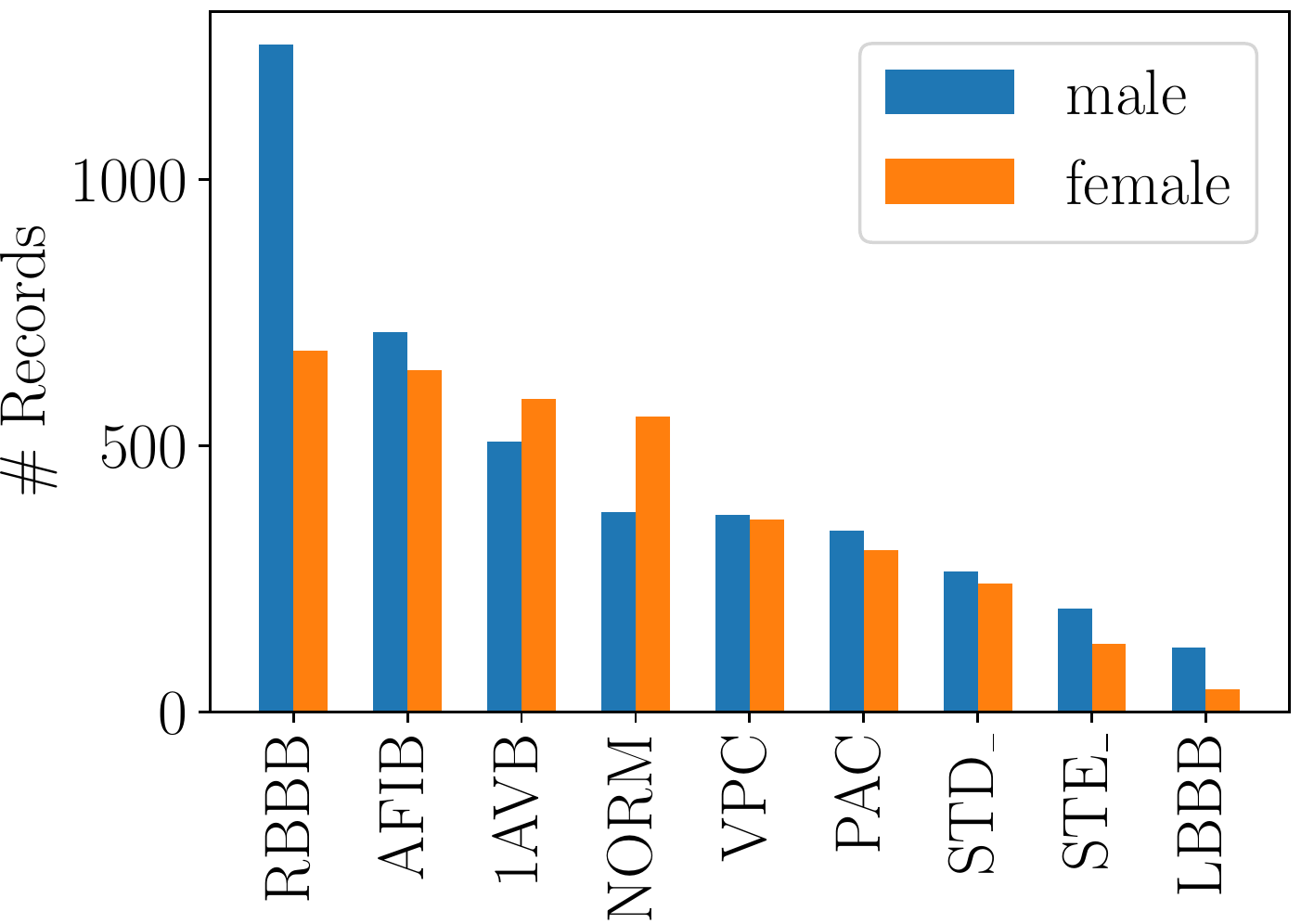}
 \caption{Summary of the ICBEB2018 dataset \cite{Liu2018ICBEB} in terms of ECG statements.}
 \label{fig:icbeb_barplot}
\end{figure}

Beyond analyses on the PTB-XL dataset itself, we see further application of it as generic pretraining resource for ECG classification task, in a similar way as ImageNet \cite{deng2009imagenet} is commonly used for pretraining image classification algorithms. 
One freely accessible dataset from the literature that is large enough to reliably quantify the effects of transfer learning is the ICBEB2018 dataset,
which is based on data released for the 1st China Physiological Signal Challenge 2018 held during the 7th International Conference on Biomedical Engineering and Biotechnology (ICBEB 2018) \cite{Liu2018ICBEB}. It comprises 6877 12-lead ECGs lasting between 6~s and 60~s. Each ECG record is annotated by up to three statements by up to three reviewers taken from a set of nine classes (one normal and eight abnormal classes, see \Cref{fig:icbeb_barplot}). We use the union of labels turning the dataset into a multi-label dataset. As the original test set is not available, we define 10 cross-validation folds by stratified sampling preserving the overall label distribution in each fold following \cite{Wagner2019PTBXL}. 

We start by analyzing the classification performance of classifiers trained on ICBEB2018 from scratch as an independent validation of the results obtained on PTB-XL. \Cref{tab:icbeb} shows the performance of classifiers that were trained using the the same experimental setup as in \Cref{sec:quantitative}. In all cases, we train a classifier from scratch by training on the first eight folds using the ninth and tenth fold as validation and test sets, respectively. Interestingly, the ICBEB2018 dataset was recently selected as training dataset for the PhysioNet/CinC challenge 2020 \footnote{\url{https://physionetchallenges.github.io/2020/}}. For this reason we also report two further label-based performance metrics that will supposedly serve as evaluation metrics in the challenge, namely a macro-averaged $F_\beta$-score ($\beta =2$) and a macro-averaged $G_\beta$-score with $\beta =2$, where $G_\beta= TP/(TP+FP+\beta \cdot FN)$, in both cases with sample weights chosen inversely proportional to the number of labels. Values of $\beta>1$ allow to assign more weight to recall than precision, which might be a desirable property. However, applying this equally to the \textit{NORM}-class seems questionable since high precision is required in this case. In addition, the corresponding scores are sensitive to the chosen classification threshold, which we determine by maximizing the $F_\beta$/$G_\beta$-score on the training set, which is an undesirable aspect as it entangles the discriminitive performance of the classification algorithm with the process of threshold determination. Nevertheless, both $F_\beta$ and $G_\beta$ show a quantitative similarity in terms of ranking between our threshold-free metrics. Comparing to the quantitative classification performance on PTB-XL as presented in \Cref{sec:quantitative}, we see a largely consistent picture on ICBEB2018 in the sense of a rather uniform performance level among the convolutional architectures, all of which remain consistent within error bars, a slightly weaker performance of the recurrent architectures and a considerable performance gap to the feature-based baseline classifier.

\begin{figure}
  \centering
    \includegraphics[width=\columnwidth]{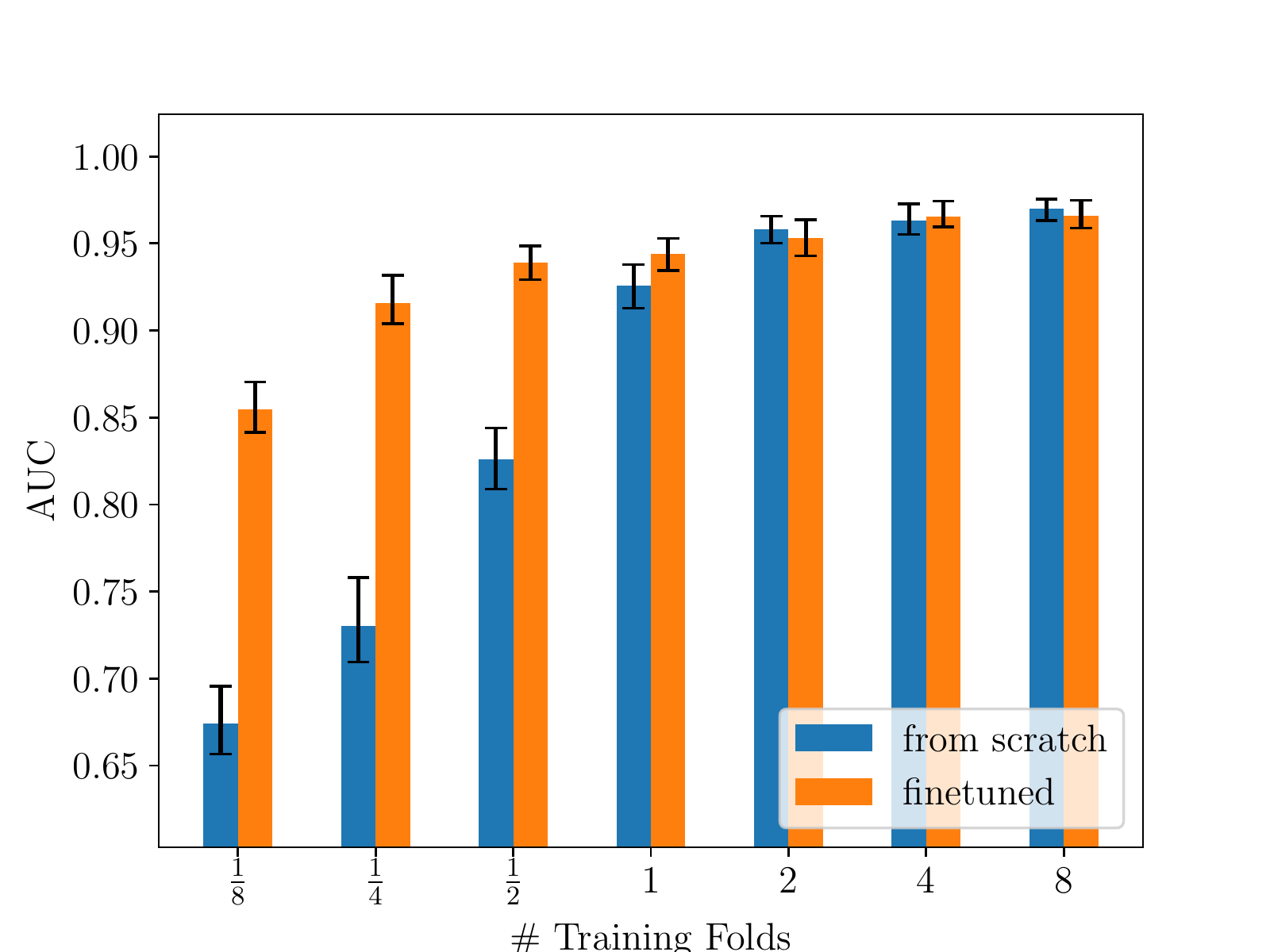}
  \caption{Effect of transfer learning from PTB-XL to ICBEB2018 upon varying the size of the ICBEB2018 training set.}
  \label{fig:transfer}
\end{figure}

In the next experiment, we leverage PTB-XL by finetuning a classifer trained on PTB-XL on ICBEB2018 data. To this end, we take a classifier trained on PTB (using \textit{all} ECG statements) and replace the top layer of the fully connected classification head to account for the different number ECG statements in ICBEB2018. This classifier is then finetuned on ICBEB2018 data.
To systematically investigate the transition into the small dataset regime, we do not only present results for finetuning on the full dataset (8 training folds) but for the full range of one eighth to eight training folds i.e.\ from 85 to 5500 training samples. For each training size and fixed model architecture (xresnet1d101), we compare models trained from scratch to models that pretrained on PTB-XL and then finetuned on ICBEB2018. \Cref{fig:transfer} summarizes the results of this experiment, and illustrates the fact that for large dataset sizes pretraining on PTB-XL does not improve the performance compared to training from scratch but even potentially slightly deteriorates it, even though the two results remain compatible within error bars. 
However, for smaller dataset sizes of a single training fold or fractions of it, we see a clear advantage from pretraining. Most notably, the performance of the finetuned model remains much more stable upon decreasing the size of the training set and consequently outperforms the model trained from scratch by a large margin in the the case of small training sizes. In the most extreme case of one eighth of the original training fold corresponds to 85 samples, where the performance of the finetuned classifier only drops by about 10\% in terms of AUC compared to a classifier trained on a training set that is 64 times larger. Since the small dataset regime is the most natural application domain for pretraining on a generic ECG dataset, we see this as a very encouraging sign for future applications of PTB-XL as a pretraining resource for relatively small datasets. 

\subsection{Age regression and gender classification}
\label{sec:exp_age_gender}
\begin{table}[t]
    \centering
    \setlength\tabcolsep{3pt}
    \caption{Age regression performance for models trained on all patients and evaluated on all/healthy/non-healthy subpopulation in terms of mean absolute error (MAE) and R-squared (R2).}
\begin{tabular}{|l|ll|ll|ll|}\hline & \multicolumn{2}{c|}{all} & \multicolumn{2}{c|}{healthy}& \multicolumn{2}{c|}{non-healthy}\\
\hline
               Method &          MAE &            R2 &          MAE &            R2 &          MAE &            R2 \\
\hline
          lstm &     7.54(22) &     .703(17) &     7.22(32) &     .688(28) &     7.78(28) &     .541(42) \\
    lstm\_bidir &     7.42(20) &     .709(22) &     7.07(31) &     .696(24) &     7.69(26) &     .550(39) \\
  xresnet1d101 &     7.35(22) &     .713(19) &     6.93(26) &     .711(25) &     7.68(27) &     .543(44) \\
 resnet1d\_wang &     7.17(18) &  \bfseries .728(20) &  \bfseries 6.86(30) &  \bfseries .721(24) &     7.41(25) &     .573(37) \\
      fcn\_wang &     7.28(23) &     .719(21) &     6.96(27) &     .712(23) &     7.54(24) &     .557(43) \\
   inception1d &  \bfseries 7.16(18) &  \bfseries .728(21) &     6.89(30) &     .715(25) &  \bfseries 7.38(21) &  \bfseries .580(38) \\
             \hline ensemble &  \bfseries 7.12(20) &  \bfseries .734(19) &  \bfseries 6.80(28) &  \bfseries .724(24) &  \bfseries 7.37(26) &  \bfseries .586(33) \\
\hline
\end{tabular}

    \label{tab:age}
\end{table}

The following experiment is inspired by the recent work from \cite{Attia2019Age} that demonstrated that deep neural networks are capable of accurately inferring age and gender from standard 12-lead ECGs. Here, we look into both tasks again based on PTB-XL. The experiment is supposed to illustrate the possibility of leveraging demographic metadata in the PTB-XL dataset.
We applied the same model architectures from \Cref{sec:quantitative} but with adjusted final layers, where for gender prediction a binary and for age prediction a linear output neuron was trained and optimized such that the binary cross-entropy or mean squared error is minimized respectively. Both networks were trained separately but with the same train-test-splits and identical hyperparameters as in previous experiments, except that for final output prediction where we computed the mean of all windows instead of the maximum (as used above). In order to study the effect of pathologies on performance for this task, in addition to all subjects we also evaluated the models only for healthy subjects and for non-healthy subjects. Here, we define the set of healthy records as the set of records with \textit{NORM} as the only diagnostic label and non-healthy as its complement. 

\begin{table}[t]
  \centering
  \setlength\tabcolsep{3pt}
  \caption{Gender prediction performance for models trained on all patients and evaluated on all/healthy/non-healthy subpopulations in terms of accuracy (acc) and area under the receiver operating curve (AUC).}
\begin{tabular}{|l|ll|ll|ll|}\hline & \multicolumn{2}{c|}{all} & \multicolumn{2}{c|}{healthy}& \multicolumn{2}{c|}{non-healthy}\\
\hline
               Method &           ACC &           AUC &           ACC &           AUC &           ACC &           AUC \\
\hline
          lstm &     .833(14) &     .911(11) &     .886(18) &     .952(11) &     .785(20) &     .874(17) \\
    lstm\_bidir &     .838(14) &     .908(12) &     .893(19) &     .954(14) &     .796(23) &     .868(17) \\
  xresnet1d101 &  \bfseries .849(14) &  \bfseries .920(10) &  \bfseries .898(19) &  \bfseries .960(10) &  \bfseries .806(17) &  \bfseries .881(17) \\
 resnet1d\_wang &     .840(14) &     .909(11) &     .895(21) &     .955(12) &     .799(19) &     .869(18) \\
      fcn\_wang &     .832(13) &     .909(11) &     .882(22) &     .949(12) &     .796(22) &     .875(18) \\
   inception1d &     .836(15) &     .916(09) &     .896(20) &     .958(12) &     .787(18) &     .876(15) \\
             \hline ensemble &     .847(15) &  \bfseries .928(09) &     .896(22) &  \bfseries .962(11) &     .801(17) &  \bfseries .894(15) \\
\hline
\end{tabular}

  \label{tab:gender}
\end{table}

The results for the age regression experiment are shown in \Cref{tab:age}. Overall, testing only on healthy subjects yielded better results in each category as compared to testing only on non-healthy or all subjects (MAE=$6.86$ compared to MAE=$7.38$ and MAE=$7.16$ respectively). These observations are in line with \cite{ball2014predicting,Attia2019Age}. Furthermore, these results are competitive to \cite{Attia2019Age}, who reported a value of MAE=$6.9$ years (R-squared = $0.7$) but with thirty times more data ($\approx $20k versus $\approx $750k samples \cite{Attia2019Age}). \Cref{tab:gender} shows the corresponding results for gender prediction. As already suggested in \cite{malik2013qt,salama2014sex} the differences between male and female are also present in ECG, which is also confirmed by our model yielding a accuracy of $84.9$\%($89.8$\%) and an AUC of $0.92$($0.96$) on all(healthy) patients. This performance level, in particular on the healthy subpopulation, is competitive with results from the literature \cite{Attia2019Age} ($90.4$\% accuracy and an AUC of $0.97$). As a final word of caution, we want to stress that the results for age and gender prediction algorithms are not directly comparable across different datasets due to different dataset distributions not only in terms of the labels themselves but also in terms of co-occurring diseases. This is apparent from the performance differences of our classifier for both subtasks when evaluated on the full dataset and on the two different subpopulations.

\subsection{Signal quality assessment}
\label{sec:sig_quality}
As part of a technical validation of the database each sample underwent a second iteration by a technical expert to annotate the data with respect to signal artifacts, see \cite{Wagner2019PTBXL} for a detailed description. The annotations were given without any regular syntax, for this reason the annotations were coded as a binary targets, where targets are set to one if any annotation is given for \textit{NOISE} (either globally present static noise  (\textit{static\_noise}) or local bursts of high voltage induced by external sources (\textit{burst\_noise})), \textit{DRIFT} (baseline wandering). In total this binary target is set for $\approx 22$\% (i.e.\ $\approx 78$\% of signals contain no artifacts). Using these annotations coded as binary targets might help to develop a signal quality classifier for creating validation data to test for robustness with respect to artifacts. For this purpose we conducted experiments along the lines of \Cref{sec:quantitative} i.e.\ again using the same models, hyper-parameters and train-test-splits as above. Overall, our models reach AUC scores around $0.81$, which seems to indicate a slightly weaker predictive performance compared to ECG statement prediction models discussed in \Cref{sec:quantitative}, even though performance measures for different tasks are obviously not directly comparable. According to a first analysis, a significant portion of this performance deficiency can be attributed label noise (i.e.\ missing annotation in case of artifacts or misleading annotation in case of normal signals). However, a more thorough analysis should attempt to incorporate the full report strings instead of just binary labels. In any case, models trained on this task can still be used as a prescreening procedure for ECG quality assessment.
\begin{figure}[htb]
  \centering
  \includegraphics[width=.9\columnwidth]{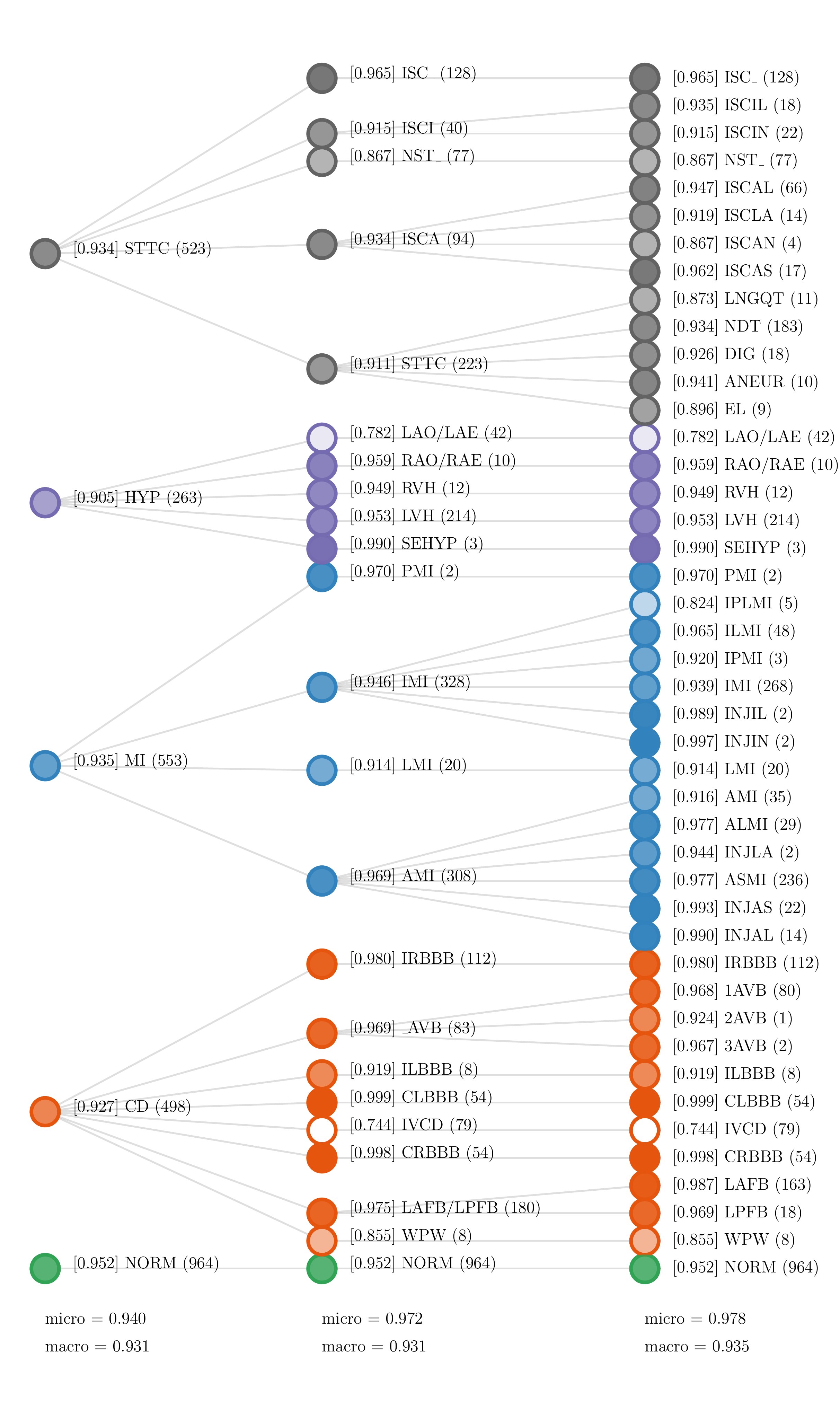}
  \caption{Hierarchical decomposition of class-specific AUCs onto subclasses and individual diagnostic statements exhibiting hidden stratification, i.e.\ inferior algorithmic performance on certain diagnostic subpopulations that remains hidden when considering only the superior superclass performance, see the description in \Cref{sec:hiddenstrat} for details. AUC is given in square brackets and the number of label occurrences in the test set in parentheses.  The transparency of each colored node is relative to the minimum and maximum AUC in the last layer.}
  \label{fig:hidden_strat}
\end{figure}

\section{Deeper insights from classification models}
Until now we investigated our experiments quantitatively in order to compare different model architectures. However, a quantitatively evaluation focusing on overall predictive performance, as presented in the previous section, might not take important qualitative aspects into account, such as the predictive performance for single, potentially sparsely populated ECG statements. Here, we focus our analysis on a single xresnet1d101 model, but we verified that the results presented below are largely consistent across different model architectures.
\label{sec:insights}
\subsection{Hierarchical organization of diagnostic labels}

As first analysis, we cover the hierarchical organization of diagnostic labels and its impact on predictive performance. The PTB-XL dataset provides proposed assignements to one of five superclasses and one of 23 subclasses for each diagnostic ECG statement, which represents one possible ontology that can be used to organize ECG statements. In \Cref{fig:hidden_strat}, we show the hierarchical decomposition for the diagnostic labels in sub- and superclasses, where we propagated predictions from experiment \textit{diag.} upwards the hierarchy over \textit{sub-diag.} to \textit{super-diag.} by summing up prediction probabilities of the corresponding child nodes and limiting the output probabilities to one. We experimented with other aggregation strategies such as using the maximum or the mean of the predictions of the child nodes but observed only minor impact on the results. The same holds for models trained on the specific level, where no propagation is needed.
The training of hierarchical classifiers is a topic with a rich history in the machine learning literature, see for example \cite{Silla2010} for a dedicated review and \cite{pmlr-v80-wehrmann18a} for a recent deep learning approach to the topic. Extensive experiments on this topic are beyond the scope of this manuscript, but our first experiments on this topic indicate that the performance of a model trained on a coarser granularity is largely compatible or in some cases even slightly inferior to a model trained on the finest label granularity and propagating prediction scores upwards the label hierarchy.

\subsection{Hidden stratification and co-occurring pathologies}
\label{sec:hiddenstrat}
\begin{figure}[t]
  \centering
  \includegraphics[width=.8\columnwidth]{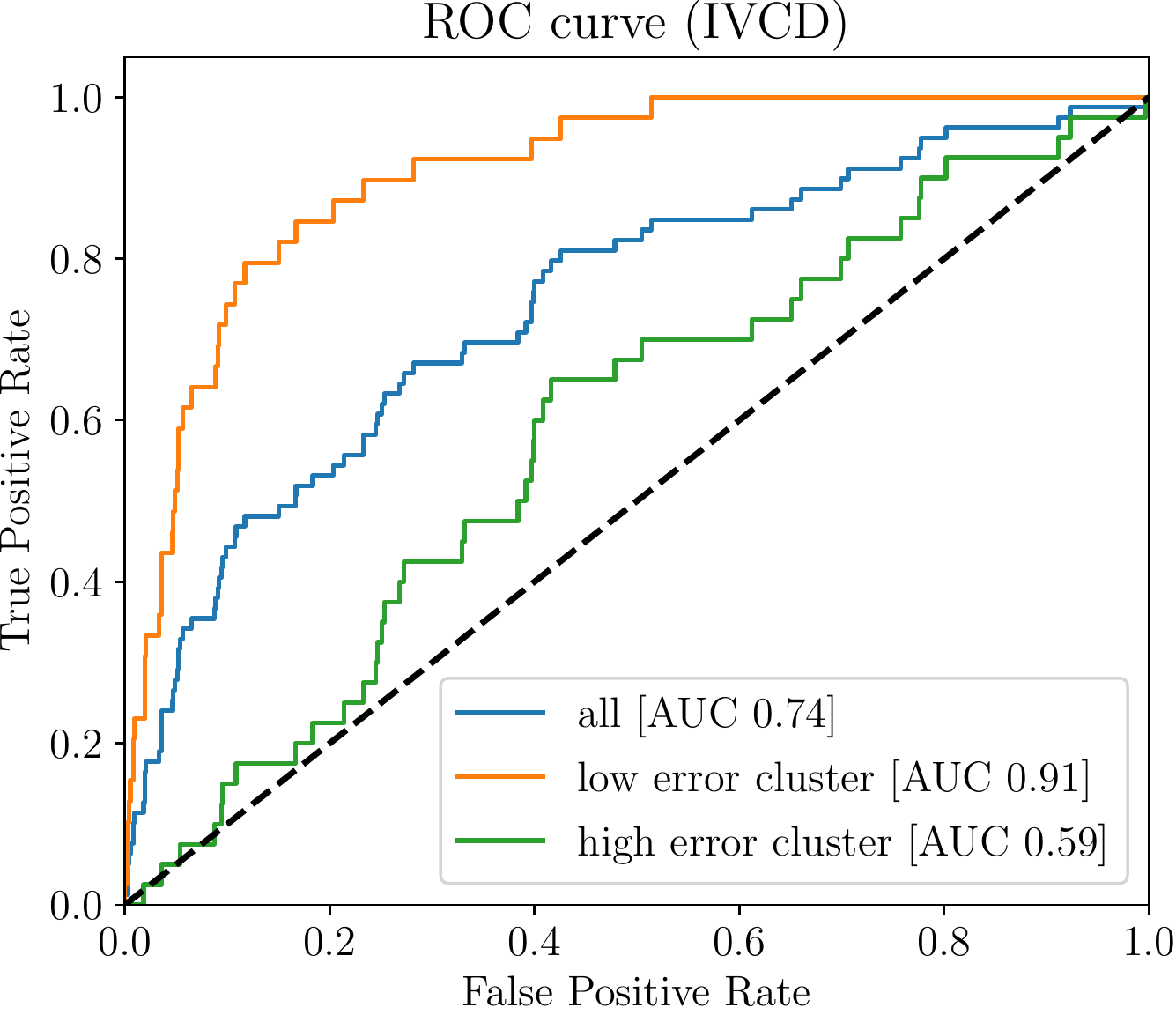}
  \caption{AUC curves for two subset of samples revealing hidden stratification within the \textit{IVCD} class.}
  \label{fig:example_hidden_strat}
\end{figure}
The hierarchical organization of the diagnostic labels allows for deeper insights and potential pitfalls of model evaluation that are crucial for clinical applications. In particular, we focus on the issue of \textit{hidden stratification} that was put forward in \cite{oakdenrayner2019hidden} and describes potential inferior algorithmic performance on certain diagnostic subpopulations that remains hidden from the outside if only the superclass performance is reported. We analyze this effect in a top-down fashion using the results obtained by propagating the finest granularity scores upwards the label hierarchy as described above. In \Cref{fig:hidden_strat}, we illustrate how the label AUC of a particular superclass or subclass decomposes into the label AUCs of the corresponding subclasses. One reason for weak classifier performance are ECG statements classes that are too scarcely populated to allow training a discriminative classifier on them and for which also the score estimate on the test set is unreliable due to the small sample size. However, there are further ECG statements that stand out from other members of the same subclass, where the performance deficiency cannot only be attributed to effects of small sample sizes. For example, consider the classes \textit{NST\_} (non-specific ST changes), \textit{LAO/LAE} (left atrial overload/enlargement) and \textit{IVCD} (non-specific intraventricular conduction disturbance (block)) in the bottom layer of the hierarchy, where the classifier shows a weak performance, which is in fact hidden when reporting only the corresponding superclass or subclass performance measures. At least for \textit{NST\_} and \textit{IVCD}, these findings can be explained by the fact that both statements are by definition non-specific ECG statements and potentially subsum rather heterogenous groups of findings.

Although identifying hidden stratification is straightforward to identify in hindsight given the hierarchical organization of the diagnostic labels, \cite{oakdenrayner2019hidden} also demonstrated how to identify groups of samples exhibiting hidden stratification for a given class label under consideration using an unsupervised clustering approach. For demonstration, we carried out such a comparable analysis for \textit{IVCD} in order to understand the comparably weak classification performance on the particular statement compared to other conduction disturbances. Indeed, clustering the model's output probabilities revealed two clusters, where one subset performed much better than the other as can be seen in \Cref{fig:example_hidden_strat}. Interestingly, it turned out that the two clusters largely align with the presence/absence of \textit{NORM} as additional ECG statement. The blue line (all) represents the performance as is (AUC 0.74), the green line is the performance for samples out of one cluster (AUC 0.59, for which most of the sample were also associated with \textit{NORM}), the orange line for the second cluster (AUC 0.91, predominantly samples without \textit{NORM}). As can be seen clearly, samples with \textit{IVCD} in combination with \textit{NORM} are much harder to classify. 

These kinds of investigations are very important for the identification of hidden stratification in the model which are induced by data and their respective labels \cite{oakdenrayner2019hidden}. Models trained on coarse labels might hide this kind of clinically relevant stratification, because of both subtle discriminative features and low prevalence. Further studies might investigate hidden stratification below our deepest level of labels. At this point, it remains to stress that the PTB-XL dataset does not provide any clinical ground truth on the considered samples but only provides cardiologists' annotations based on the ECG signal itself, which could compromise the analysis. However, we still see an in-depth study towards the identification subgroups with certain combinations of co-occurring ECG statements/pathologies, along the lines of the example of \textit{IVCD} presented above, as a promising direction for future research in the sense that it can potentially provide pointers for future clinical investigations.

\begin{figure}[t]
  \centering
    \includegraphics[width=.8\columnwidth]{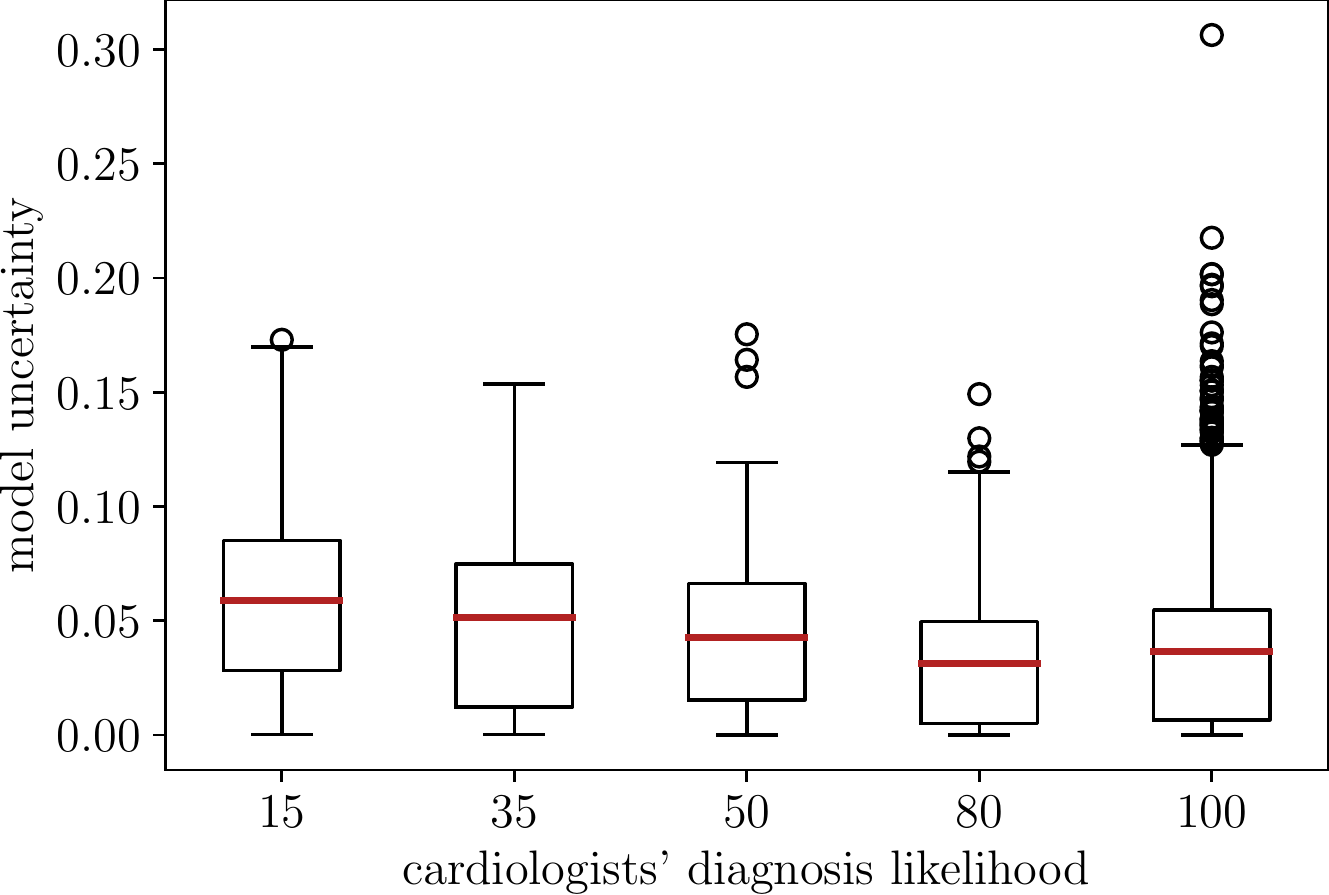}
  \caption{Relation between model uncertainty (standard deviation of ensemble predictions as in \cite{lakshminarayanan2017simple}) and diagnosis likelihood as quantified by the annotating cardiologist, see \Cref{sec:uncertainty} for details.}
  \label{fig:unc_conf}
\end{figure}

\begin{figure*}[!ht]
  \centering
  \begin{subfigure}[b]{0.45\textwidth}
      \includegraphics[width=\textwidth]{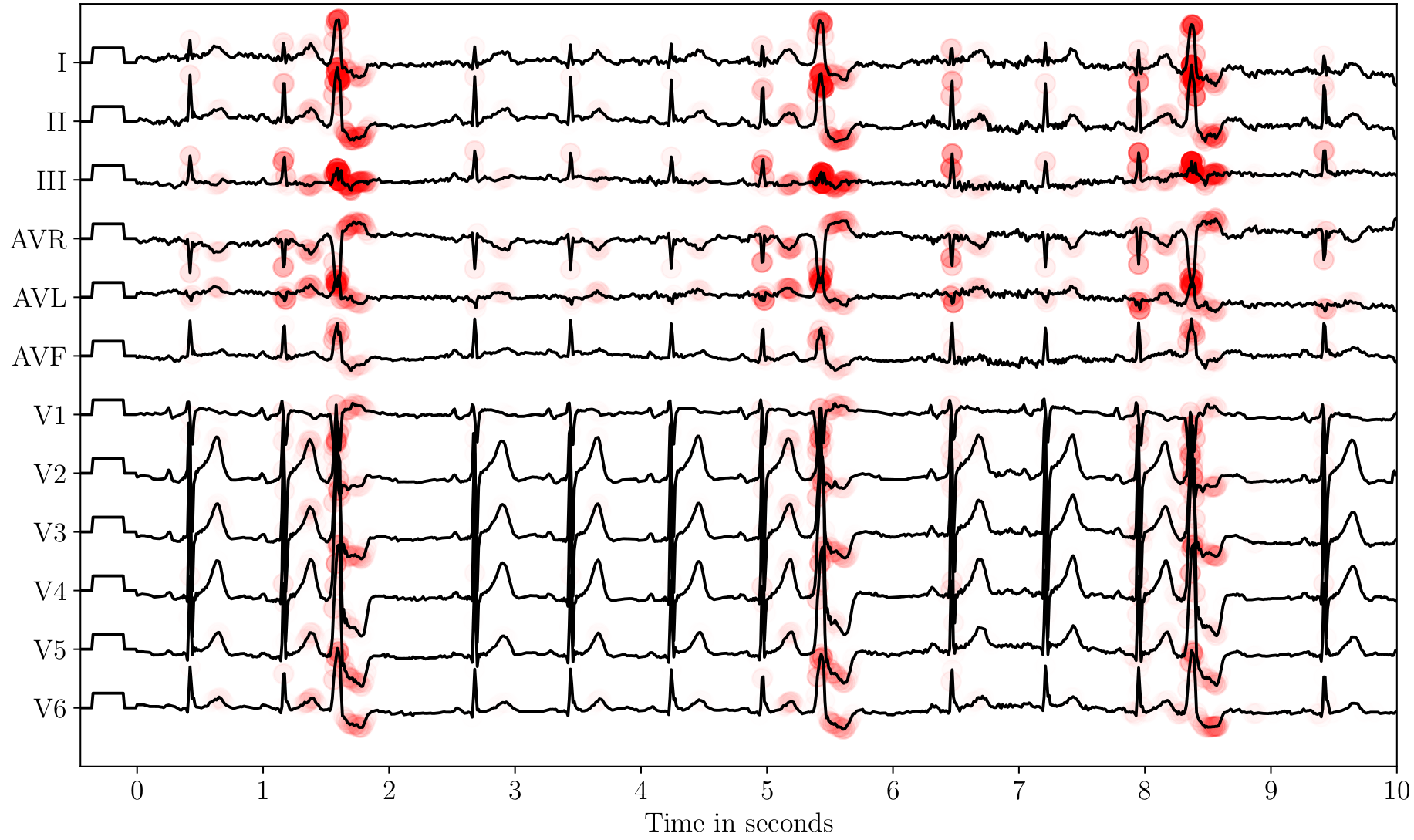}
      \caption{PVC}
      \label{fig:explain_pvc}
  \end{subfigure}
  \begin{subfigure}[b]{0.45\textwidth}
    \includegraphics[width=\textwidth]{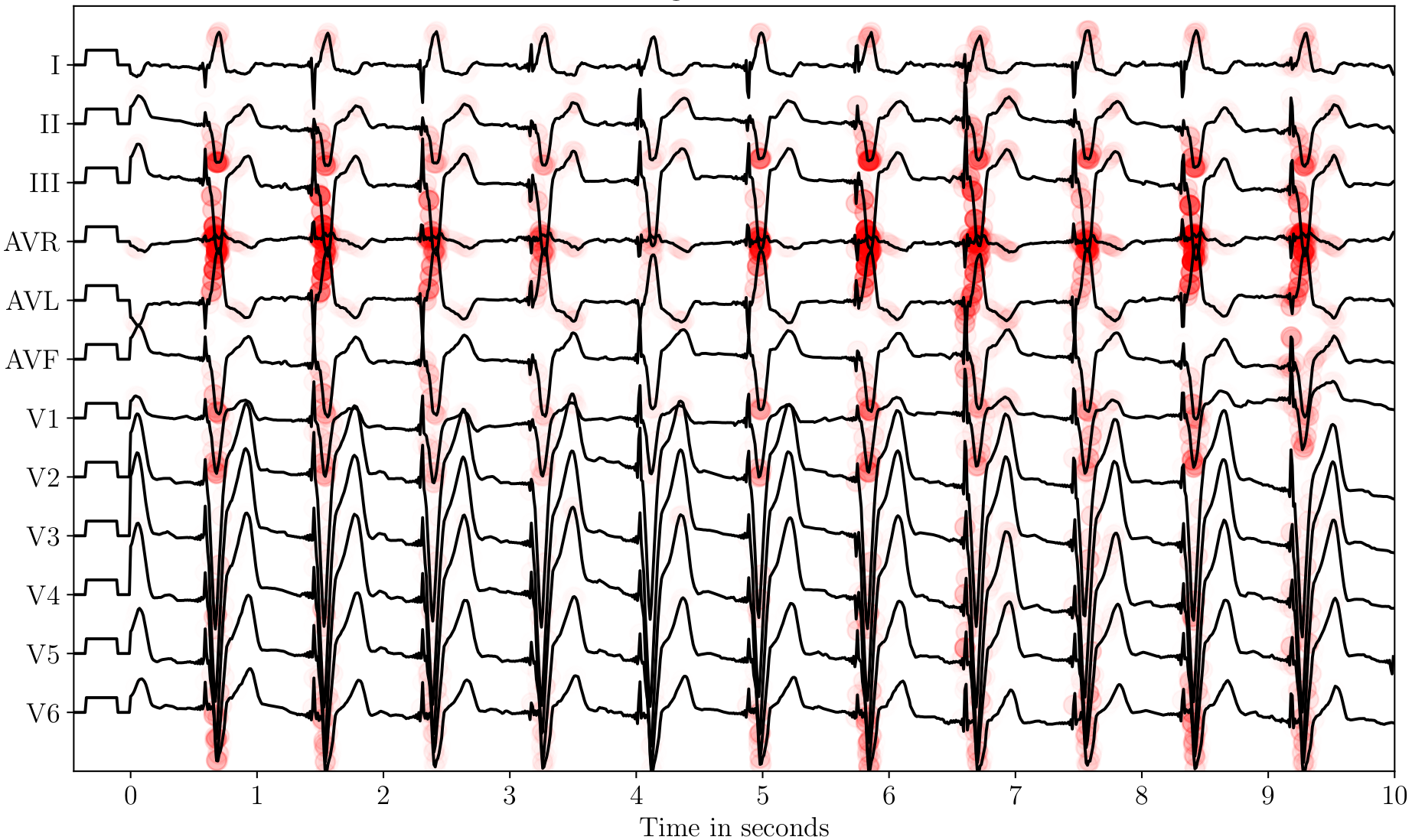}
    \caption{PACE}
    \label{fig:explain_pace}
  \end{subfigure}
\caption{Two exemplary attribution maps for a resnet model for the classes PVC (left) and PACE (right).}
\label{fig:explain}
\end{figure*}
\subsection{Model uncertainty and diagnosis likelihoods}
\label{sec:uncertainty}
Besides this hierarchical organization of diagnostic labels, PTB-XL comes along with associated likelihoods for each diagnostic label ranging from 15 to 100, where 15 indicates less and 100 strong confidence for one label. These likelihoods were extracted from the original ECG report string for all diagnostic statements based on certain keywords \cite{Wagner2019PTBXL}. As an initial experiment to assess the quality of this likelihood information, we compare the likelihoods to model uncertainty estimates for a model trained on diagnostic statements. To quantify the model uncertainty, we follow the simple yet very powerful approach put forward in \cite{lakshminarayanan2017simple} that defines model uncertainty via the variance of an ensemble of identical models for different random initializations. Here, we use an ensemble of 10 models and for simplicity even omit the optional stabilizing adversarial training step, which was reported to lead to slightly improved uncertainty estimates \cite{lakshminarayanan2017simple}, in this first exploratory analysis. In \Cref{fig:unc_conf}, we plot model uncertainty versus diagnosis likelihood and observe the expected monotonic behavior. Only the likelihood 100 stands out from this trend and shows a large number of outliers. One possible explanation for this observation is an overconfidence of human annotators when it comes to seemingly very obvious statements that goes in with the human inability to precisely quantify uncertainties, which is a well-known phenomenon in cognitive psychology, see e.g.\ \cite{windschitl1996measuring}. However, we perceive the overall alignment of diagnosis likelihood with model uncertainty as an interesting observation as it correlates perceived human uncertainty with algorithmic uncertainty, a statement that is normally impossible for clinical datasets due to the unavailability of appropriate labels.

\subsection{Prospects of interpretability methods}
The acceptance of machine learning and in particular deep learning algorithms in the clinical context is often limited by the fact that data-driven algorithms are perceived as black boxes by doctors. In this direction, the recent advances in the field of explainable AI has the prospect to at least partially alleviate this issue. In particular, we consider post-hoc interpretability that can be applied for a trained model, see e.g.\ \cite{xai2019}. The general applicability of interpretability methods to multivariate timeseries and in particular ECG data was demonstrated in \cite{strodthoff2018detecting}, see also \cite{westhuizen2017techniques,vijayarangan2020interpreting} for futher accounts on interpretability methods for ECG data. Here, we focus on exemplary for the form statement ``premature ventricular complex'' (\textit{PVC}) and the rhythm statement \textit{PACE} indicating an active pacemaker. The main reason for choosing these particular classes is the easy verifiable also for non-cardiologists. In \Cref{fig:explain}, we show two exemplary but representative attribution maps obtained via the $\epsilon$-rule with $\epsilon=0.1$ within the framework of layer-wise relevance propagation \cite{bach2015pixel}. For \textit{PVC} the relevance is located at the extra systole across all leads. For \textit{PACE}, the relevance is scattered across the whole signal aligning nicely with the characteristic pacemaker spikes (just before each QRS complex) in each beat. It is a non-trivial finding that the relevance patterns for the two ECG statements from above align with medical knowledge. A more extensive, statistical analysis of the attribution maps both within patients across different beats and across different ECGs with common pathologies is a promising direction for future work. 
\section{Summary and conclusions} 
Electrocardiography is among the most common diagnostic procedures carried out in hospitals and doctor's offices. We envision a lot potential for automatic ECG interpretation algorithms in different medical application domains, but we see the current progress in the field hampered by the lack of appropriate benchmarking datasets and well-defined evaluation procedures. We propose a variety of benchmarking tasks based on the PTB-XL dataset \cite{Wagner2019PTBXL} and put forward first baseline results for deep-learning-based time classification algorithms that are supposed to guide future reasearchers working on this dataset. We find that convolutional, in particular resnet- and inception-based, architectures show the best performance but recurrent architectures are also competitive for most prediction tasks. Furthermore, we demonstrate the prospects of transfer learning by finetuning a classifier pretrained on PTB-XL on a different target dataset, which turns out to be particularly effective in the small dataset regime. Finally, we provide different directions for further in-depth studies on the dataset ranging from the analysis of co-occurring pathologies, over the correlation of human-provided diagnosis likelihoods with model uncertainties to the application of interpretability methods.
We release the training and evaluation code for all ECG statement prediction tasks, trained models as well as the complete model predictions in an accompanying code repository \cite{coderepo}.

\bibliographystyle{IEEEtran}
\bibliography{bibfile}

\appendices
\section{Experimental details}
\label{appendix:details}
In general, our implementations follow the implementations of the architectures described in the original publications and reference implementations as closely as possible. The most significant modification in our implementations is the use of a concat-pooling layer \cite{howard2018fastai} as pooling layer, which aggregates the result of a global average pooling layer and a max pooling layer along the feature dimension. For resnets we enlarge the kernel sizes as this slightly improved the performance, consistent with observations in the literature \cite{wang2017time,fawaz2019dreamtime}. All convolutional models then use the same fully connected classification head with a single hidden layer with $128$ hidden units, batch normalization and dropout of 0.25 and 0.5 at the first/second fully connected layer, respectively. For recurrent neural networks we use concat pooling as in \cite{howard2018universal}. For reference, we typically report the performance of both unidirectional and bidirectional recurrent models, in our case LSTMs/GRUs with two layers and $256$ hidden units. As we are dealing with a multi-label classification problem, we optimize binary cross-entropy. We use 1-cycle learning rate scheduling during training \cite{smith2018disciplined} and the AdamW optimizer \cite{Loshchilov2018FixingWD}. During finetuning a pretrained classifier for transfer learning from PTB-XL to ICBEB2018, we use gradual unfreezing and discriminative learning rates \cite{howard2018fastai,howard2018universal} to avoid catastrophic forgetting i.e.\ overwriting information captured during the initial training phase on PTB-XL. 
Deep-learning models were implemented using PyTorch \cite{PytorchNIPS2019}, fast.ai \cite{howard2018fastai} and Keras \cite{chollet2015keras}. We release our implementations in the accompanying code repository \cite{coderepo}.

During training, we follow the sliding window approach that is commonly used in time series classification, see e.g.\ \cite{cui2016multi,schirrmeister2017deep,strodthoff2018detecting,yannick2019deep}. Here, the classifier is trained on random segments of fixed length taken from the full record. This allows to easily incorporate records of different length (as it is the case for ICBEB2018) and effectively serves as data augmentation. During test time, we use test time augmentation. This means we divide the record into segments of the given window size that overlap by half of the window size and obtain model predictions for each of the segments. These predictions are then aggregated using the element-wise maximum (or mean in case of age and gender prediction) in order to produce a single prediction for the whole sample. This procedure considerably increases the overall performance compared to the performance on random sliding windows without any aggregation. If not mentioned otherwise, we use a fixed window size of 2.5 seconds.

Besides our end-to-end trainable models we also compare to classic machine learning models, where a classifier is trained on precomputed statistical features such as wavelets. Here we loosely follow \cite{Sharma2017} and train a classifier on wavelet features. More specifically, we compute a multilevel 1d discrete wavelet transform (Daubechies \texttt{db4}) for each lead independently leveraging the implementation from \cite{Lee2019}. From the resulting coefficients we compute a variety of statistical features such as entropy, 5\%, 25\%, 75\% and 95\% percentiles, median, mean, standard deviation, variance, root of squared means, number of zero and mean crossings. Different from \cite{Sharma2017}, the features were then used to train a shallow neural network with a single hidden layer in order to be able to address multi-label classification problems with a large number of classes in a straightforward manner. Note that the classifier from \cite{Sharma2017} included a number of additional features and preprocessing steps and might therefore lead to an improved score compared to our implementation.

Again following the example of the CAFA challenge, we provide 95\% confidence intervals via empirical bootstrapping on the test set, in our case with 1,000 iterations. More specifically, we report the point estimate from evaluating on the whole test set and estimate lower and upper confidence intervals using the bootstrap examples. In summary tables, we typically report only the point estimate and the maximal absolute deviation between point estimate and lower and upper bound, where for example $0.743(09)$ is supposed to be understood as $0.743\pm 0.009$. We deliberately decided not to exclude sparsely populated classes from the evaluation. Due to the stratified sampling procedure underlying the fold assignments in \cite{Wagner2019PTBXL} point estimates can evaluated for all metrics. However, during the bootstrap process it is not guaranteed that at least one positive sample for each class is contained in each bootstrap sample. In such a case, metrics such as the term-centric macro-AUC cannot be evaluated. To circumvent this issue, we discard such bootstrap samples and redraw until we find at least one positive sample for each class. For metrics such as sample-centric Fmax that can be evaluated without any constraints on the bootstrap samples, we verified empirically that this procedure only marginally impacts the corresponding confidence intervals. For later reference, we store the selection of bootstrap samples, evaluate the confidence intervals for all algorithms on this fixed set of samples and provide this as part of our code repository \cite{coderepo}.

\section{Train-test splits and sampling frequency}
\label{sec:misconceptions}

In this section, we investigate the impact of two crucial experimental parameters on the classification performance, namely the effect of using random splits disregarding patient assignments and using 500 Hz compared to 100 Hz data as input.

Concerning the first aspect, we noticed that many literature approaches in the field ECG analysis perform train-test splits using random splits on individual ECGs or even individual beats rather than patients. This leads to a systematic overestimation of generalization performance, since during prediction on the test set the model can exploit training data for patients with multiple ECGs having both samples in train and test split. We substantiate this claim by comparing the results from \Cref{sec:quantitative} with models trained on random splits for the experiment \textit{super-diag.}. By random we refer to random splits based on ECGs rather than patients but with the same  splitting procedure i.e.\ stratified sampling in order to ensure balanced label distributions. And even more importantly, we also maintain two \textit{clean} folds for validation and testing i.e.\ only samples where \textit{validated\_by\_human} is set to true. This point is important since splitting disregarding clean folds yields even better results, which might be attributed to slight mismatches in the distribution of the clean folds and the remaining training folds.
\Cref{tab:summary_misconceptions} shows the results for this experiment, where the overestimation becomes apparent when comparing the ensemble model for example where in terms of both metrics the performance is increased only by this effect. This is a strong indication for our claim that most of results reported in previous literature overestimates generalization performance due to a data leakage arising from having one patient in both training and test set. The effect is even more pronounced for datasets where the fraction of test set samples of patients with records already contained in the training set is larger, as observed by \cite{Sharma2017} for the original PTB diagnostic ECG dataset \cite{bousseljot1995Oeff}. We tested only those samples where the leakage was most severe (i.e.\ patients having most ECGs in training data) and observed a significant gain in performance and gap in loss confirming our claim. 

\begin{table}[t]
    \centering
    \setlength\tabcolsep{3pt}
    \caption{Investigating impact of random train-test splits (disregarding patient assignments) and increasing the temporal resolution (sampling frequency of 500~Hz) compared to the setup used throught this article (train-test splits considering patient assignments and a sampling frequency of 100~Hz).}
\begin{tabular}{|l|ll|ll|ll|}
\toprule
  & \multicolumn{2}{c|}{strat. \& 100Hz}& \multicolumn{2}{c|}{strat. 500Hz}& \multicolumn{2}{c|}{rnd. 100Hz}\\ Method & AUC & Fmax & AUC & Fmax & AUC & Fmax \\
\midrule
 fcn\_wang & .925(06) & .817(12) & .919(06) & .806(10) & .931(05) & .812(11) \\
 inception1d & .921(06) & .810(11) & .931(05) & \bfseries .824(10) & .920(05) & .801(11) \\
 lstm & .927(05) & .820(09) & .922(06) & .809(11) & .931(06) & .814(11) \\
 resnet1d\_wang & \bfseries .930(06) & \bfseries .823(10) & .922(06) & .809(09) & \bfseries .934(06) & \bfseries .821(09) \\
 Wavelet+NN & .874(08) & .734(11) & .852(08) & .709(12) & .877(08) & .735(12) \\
 lstm\_bidir & .921(06) & .815(10) & .919(06) & .807(13) & .925(06) & .806(11) \\
 xresnet1d101 & .928(06) & .815(12) & \bfseries .933(06) & .821(11) & .929(07) & .807(10) \\
 \hline naive & .500(00) & .448(10) & .500(00) & .448(12) & .500(00) & .452(17) \\
 \hline ensemble & \itshape \bfseries .934(05) & \itshape \bfseries .825(12) & \itshape .931(05) & \itshape .818(11) & \itshape \bfseries .937(05) & \itshape .819(11) \\
\bottomrule
\end{tabular}

    \label{tab:summary_misconceptions}
  \end{table}

Finally, we investigate the prospects of using input data at a higher sampling frequency of 500~Hz compared to 100~Hz that was used throught the rest of the manuscript. To investigate this claim, we applied the same pipeline as described in \Cref{sec:quantitative} to input data sampled at 500~Hz. Also our inputs to the networks were adjusted such that each window consists of five seconds, this results in 2500 timesteps (as compared to 500 timesteps for 100 Hz). In order to compensate for the higher sampling frequency we also increased the filter size of convolutional filters by a factor of five, while the rest of hyperparameters (number of filter, layers etc.) were left unchanged compared to the case of 100~Hz. In our experiments, we found no compelling evidence for this apprehension, i.e.\ no significant gain in performance in all our metrics for diagnostic tasks, as can be seen in \Cref{tab:summary_misconceptions}. This observation is in accordance with \cite{kwon2018electrocardiogram}, where the authors also came to the conclusion that 100 Hz is still sufficient for models operating in time domain but not for models in frequency domain. For this reason it is might not even be a fair comparison to compare our baseline Wavelet model for 100 Hz as it was done in \Cref{sec:quantitative}. Although this seems plausible, additional considerations might be necessary to compare this issue in a more reliable way. Nevertheless we believe that the gain in performance will be negligible for our models, since there are more obvious issues affecting performance, e.g.\ dealing with all sorts of artifacts and label noise, preprocessing and more resilient and effective training procedures.

\end{document}